\documentclass[runningheads]{llncs}
\usepackage{graphicx}
\usepackage{amsmath}
\begin{document}
\title{Joint Dermatological Lesion Classification and Confidence Modeling with Uncertainty Estimation}
\titlerunning{Lesion Classification and Confidence Modeling with Uncertainty Estimation}

%

\author{Gun-Hee Lee\inst{1} \and
Han-Bin Ko\inst{2} \and
Seong-Whan Lee\inst{1}}
\authorrunning{G. H. Lee et al.}
%
\institute{Korea University, Seoul, Republic of Korea\\
\email{\{gunhlee, sw.lee\}@korea.ac.kr}
\and
VUNO Inc., Seoul, Republic of Korea\\
\email{hanbin.ko@vuno.co}}

\maketitle              
\begin{abstract}
Deep learning has played a major role in the interpretation of dermoscopic images for detecting skin defects and abnormalities. However, current deep learning solutions for dermatological lesion analysis are typically limited in providing probabilistic predictions which highlights the importance of concerning uncertainties. This concept of uncertainty can provide a confidence level for each feature which prevents overconfident predictions with poor generalization on unseen data. In this paper, we propose an overall framework that jointly considers dermatological classification and uncertainty estimation together. The estimated confidence of each feature to avoid uncertain feature and undesirable shift, which are caused by environmental difference of input image, in the latent space is pooled from confidence network. Our qualitative results show that modeling uncertainties not only helps to quantify model confidence for each prediction but also helps classification layers to focus on confident features, therefore, improving the accuracy for dermatological lesion classification. We demonstrate the potential of the proposed approach in two state-of-the-art dermoscopic datasets (ISIC 2018 and ISIC 2019).

\end{abstract}

\section{Introduction}
Deep learning has remarkably developed and improved the ares of research, e.g. computer vision \cite{pf1,pf2,pf3,pf4,pf5}, natural language processing, data analysis, signal processing, etc. They even improve the state-of-the-art performances in skin lesion analysis, including lesion classification \cite{Derm2}, localization \cite{Der2}, and segmentation \cite{um2}. As classification of dermatological lesion received attention, the International Skin Imaging Collaboration (ISIC) \cite{isic} developed digital imaging standards for skin cancer imaging with creating a public archive containing revolutionary amount of publicly available collection of quality controlled dermoscopic images of skin lesions. The dermascopic skin lesion dataset \cite{Dat1} which contains 129,450 images of skin lesion, brought a massive improvement in the field where many researchers competed to achieve state-of-the art performance. Convolutional Neural Networks (CNNs) have been widely used for melanoma classification  \cite{Derm1,Derm2} where the standard for training melanoma classification models was to tune a pretrained CNN and use ensembles. Along with the development of neural network, deep learning models started to achieve similar or better performance compared to expert dermatologists \cite{Derm2}. The accuracy measures are even now consistently improving with applying various methodologies from the computer vision fields such as data augmentation \cite{Derm_Aug} or ensembles \cite{Derm_ens}.

Although there have been many approaches, there are only few works on estimating the uncertainties of the model which is crucial for medical studies where the ground truth is uncertain by nature. Skin lesion datasets contain large variability which is concerned as high degree noise. These noises are mainly due to the image quality or other disturbing objects such as hair or lump, which distracts the model from letting out correct predictions. In addition, it is also very important to detect samples that deviate from the distribution of the training data. This is why modeling uncertainties along with results is crucial to help machine learning-based methods to interact with experts by showing how confident the model is for its prediction.

Using the obtained uncertainty to model the confidence of the predictions has been a long-standing challenge for computer vision. In this work, our idea started from this following aspect. When experts are asked to speak out the results for a medical image, they could give the result with confidence, or with ambiguity. They would list all the features which back up his or her theory, each with its own confidence level. Thus, if any feature is ambiguous, doctors will ignore the corresponding features and focus on the confident ones. This can go the same with the medical models. A good model will express the confidence for every features they give out. These can help the decision layers to know what features they need to focus on.

However, past models give a deterministic point representation for each dermascopic image in the embedding space \cite{Derm1,Derm2,Derm_Aug}. This can result in bringing sophisticated problems where each feature would be represented with overconfidence which lacks generalization on unseen data \cite{UncMed1}. As some uncertain images contain ambiguous medical features, a large shift in the embedded points is inevitable leading to false classification.

To address the above problems, we propose an uncertainty-aware classification network, which jointly classifies dermatological lesion and models the confidence for its prediction. Our model addresses dermatological classification in a three-fold way: (1) The model encodes a probabilistic distribution in the latent space rather than just a point vector. (2) During the classification step, our model pays more attention to confident features while penalizing uncertain ones. (3) Along with the prediction, the model uses the estimated uncertainty value to model the confidence for the prediction.

\begin{figure*}[t]
    \begin{center}
    \includegraphics[width=1.0\linewidth]{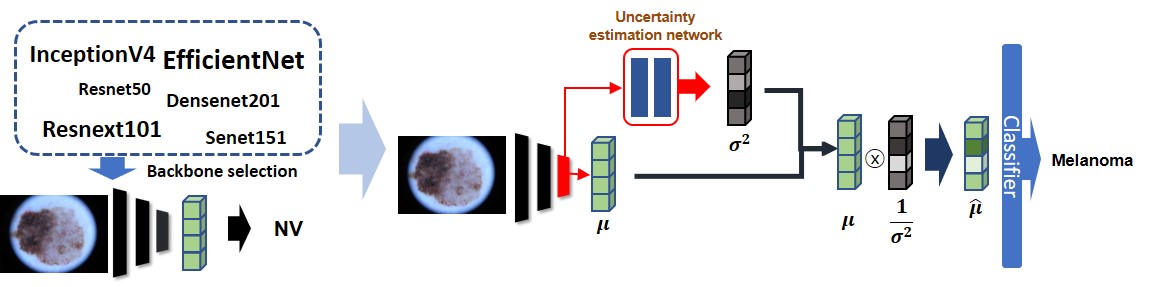}
    \end{center}
    \caption{The flow of dermatological image classification network with uncertainty estimation. We first choose the backbone  structure from various kinds of base models. Our model estimates a probabilistic distribution of the latent features. We use the inverse of estimated variance as a weight for the features which can help the decision layers to focus on confident features}
    \label{fig:fig3}
\end{figure*}
\section{Related Work}
\textbf{Dermascopic Image Classification.} In this section, we briefly review the literature in model design for skin lesion classification. We focus on CNN models which are the state of the art in the image classification field. For past days, Resnet \cite{Res}, Densenet \cite{Den} and Inception-v3 \cite{Inc} architectures were widely used as a base model. \cite{tone} proposes a CNN model which performed on par with 145 dermatologists for melanoma classification. \cite{ttwo} uses the fusion of deep learning and handcrafted method for higher diagnostic accuracy on melanoma classification. On ISIC classification challenge showed a much wider range of architectures, including Squeeze-and-Excitation Networks \cite{Sen}, Dual Path Networks \cite{Dua}, PNASNet \cite{Pna}, EfficientNet \cite{Eff}, etc. As the CNN architecture has proved to achieve excellent performance on these tasks, \cite{Derm_Aug} uses data augmentation, and \cite{Derm_ens} uses ensembles to further improve the accuracy. However, as mentioned in section I, all of the above models use deterministic embeddings, which predicts overconfident estimation of features even for the unseen or uncertain data.

\noindent\textbf{Uncertainty Learning in DNNs.} To improve the reliability of Deep Neural Networks (DNNs), uncertainty learning is receiving more attention in a lot of fields \cite{Unc1,Unc2}. The concept of uncertainty is generally divided into two uncertainties, epistemic and aleatoric. Model uncertainty or Epistemic uncertainty refers to the uncertainty of model parameters given the training data. This uncertainy arises from the insufficient training data which lacks to correctly infer the underlying data-generating function. Data uncertainty or aleatoric uncertainty measures the noise inherent in the observation. Unlike epistemic uncertainty, aleatoric uncertainty arises due to the variables or measurement errors which cannot be decreased by obtaining more data. For computer vision tasks, uncertainties are mainly modeled with Bayesian analysis. The uncertainties are formed as a probabilistic distribution for the model parameters or model inputs. Since Bayesian inference is complex to apply directly to neural networks, approximation techniques such as Laplace approximation, Markov chain Monte Carlo methods, and variational inference methods are widely used. However, there is not much work for estimating the uncertainties in the medical domain compared to other tasks \cite{Unc2,UncMed1,Unc3}, especially for dermatological images. Also, adopting Bayesian analysis or approximations techniques to recent models is still complex. This is why we propose an non-complex way to estimate epistemic uncertainty which can be used along with any other existing models. We also investigate the usage of uncertainty in dermatological lesion classification where we can both improve the accuracy and model prediction confidence.
\section{Method}
This paper proposes a network for joint dermatological lesion classification and confidence modeling with uncertainty estimation. Our model as in Fig. \ref{fig:fig3} represents features as a probabilistic Gaussian distribution for the given lesion image. The mean of the distribution is the most likely feature value which is estimated by a base model. The variances for each latent feature are estimated by an additional uncertainty estimation network. We estimate the confidence as the inverse of the variance for each feature. The measured confidence is then pooled with every corresponding feature before it is passed to the decision layers. This way, the decision layers will focus more on the confident features and ignore the uncertain ones. In this section, we explain how we estimate the probabilistic Gaussian distribution for the latent features and use the inverse of the variance as a weight of each feature. Then, we finally will show how we use the measured uncertainty to model the confidence value of each predictions.

\subsection{Mean Estimation}

For the mean value of the distribution we first choose a backbone network and then estimate a deterministic feature by training the CNN model. Note that, backbone networks which are commonly used in other skin-analysis researches or competitions were selected for training. Given the input image, we apply a weighted cross entropy loss as follows:

\begin{equation}
        L_{w c e}=-\sum_{c=1}^{C} n_{c} t_{c} \log \left(p_{c}\right).
\label{E1}
\end{equation}

The weight $n_{c}=(N/N_{c})^{k}$ balances training for each class. Here, $N$ is the total number of the dataset, $N_{c}$ is the number of data for class $c$ and $k$ is the factor that controls the balance severity. Also, $C$ represents the total class number, $t_{c}$ to a one-hot code vector for the corresponding class $c$, and $p_{c}$ to the softmax value of the score for class $c$.

\subsection{Variance Estimation}

As mentioned earlier, our goal is to estimate a probabilistic Gaussian distribution for the latent representations which can be expressed as follows:

\begin{equation}
        p\left(\mathbf{z} | \mathbf{x}\right) = \mathcal{N}\left(\mathbf{z} ; {\mu}, {\sigma}^{2} \right),
\label{E1}
\end{equation}
where $\mu$ and $\sigma^{2}$ are D-dimensional vectors predicted by each network from the input image $x$. The $\mu$ value, which is estimated from the base network, should encode the most likely lesion features of the input image. The $\sigma$ value is estimated by an additional uncertainty estimation network which measures model's variance along each feature dimension.

An ideal embedding space $Z$ should only encode lesion-related features where each lesion category should have a unique embedding $\mathbf{z} \in Z$ that best represents the lesion class. An ideal model would focus to estimate the same $z$ for every image in the same lesion class. For the representation pair of images ($x_{i}$, $x_{j}$), we estimate the likelihood $\mathrm{S}\left(z_{i}, z_{j}\right)$ of them belonging to the same lesion class with also sharing the same Gaussian distribution: $p(\Delta z=0)$, for two latent vectors $z_{i}  \sim p($x$ \mid \mathbf{z_{i}})$ and $z_{j} \sim p($x$ \mid \mathbf{z_{j}})$. Specifically, the equation can be represented as follows:
\begin{equation}
        \mathrm{S}\left(z_{i}, z_{j}\right)=\int p\left(\mathbf{z}_{i} | \mathbf{x}_{i}\right) p\left(\mathbf{z}_{j} | \mathbf{x}_{j}\right) \delta\left(\mathbf{z}_{i}-\mathbf{z}_{j}\right) d \mathbf{z}_{i} d \mathbf{z}_{j}.
\label{E2}
\end{equation}
We consider an alternative vector $\Delta z= z_{i} - z_{j}$, instead of directly solving the integral, where $z_{i}  \sim p($x$ \mid \mathbf{z_{i}})$ and $z_{j} \sim p($x$ \mid \mathbf{z_{j}})$. Then, $p(z_{i}=z_{j})$ is equivalent to the density value of $p(\Delta z = 0)$. The $l^{t h}$ component of $\Delta z, \Delta z^{(l)}$, can be represented as the subtraction of two Gaussian variables, which can be approximated as another normal distribution as follows:

\begin{equation}
    \Delta z^{(l)} \sim N\left(\mu_{i}^{(l)}-\mu_{j}^{(l)}, \sigma_{i}^{2(l)}+\sigma_{i}^{2(l)}\right).
    \label{E3}
\end{equation}

So then, we use log likelihood along with Gaussian approximation which solution is given as follows:

\begin{equation}
    \begin{split}
        \mathrm{R}\left(z_{i}, z_{j}\right) &= \ln \left(\mathrm{S}\left(z_{i}, z_{j}\right)\right)\\ &= 
        -\frac{1}{2} \sum_{l=1}^{D}\left(\frac{\left(\mu_{i}^{(l)}-\mu_{j}^{(l)}\right)^{2}}{\sigma_{i}^{2(l)}+\sigma_{j}^{2(l)}}+\ln \left(\sigma_{i}^{2(l)}+\sigma_{j}^{2(l)}\right)\right)-\frac{D}{2} \log 2 \pi,
         \label{E4}
    \end{split}
\end{equation}
where $D$ is the total number of dimensions for the latent vector. We use the above equation as a loss to train the uncertainty estimation network, which will ideally output the correct variance value for each latent feature.


\subsection{Confidence Pooling}

We use the estimated variance value as weights for each feature to make a new latent feature. We define confidence value $q_{n}$ as a normalized value in range of (0,1] for $c_{n} = {1/\sigma_{n}^{2}}$, where $\sigma^{2}$ is the output of the uncertainty network and $n$ represents the $n^{th}$ dimension. Then we can obtain a new mean value for all $D$ dimensions as follows:

\begin{equation}
     \hat{{\mu}}_{n}=\frac{{q_{n}\mu}_{n}}{\sum_{j}^{D} q_{j}}.
     \label{E6}
\end{equation}

This metric $\hat{\mu}_{j}$ denotes the new mean value pooled by its own variance. The confidence pooling method helps classification layer to focus on confident features and ignore the uncertain ones. The concept of confidence pooling is different from attention mechanisms where attention mechanisms pay more attention to parts of the input while confidence pooling pays attention to the features with more confidence. This way, the network can be robust to noises or uncertainties in the input.

\subsection{Confidence Modeling} \label{cm}

We measure confidence of the feature by using the variance estimated from the above section. As we have the value of weights for the classification layer, the score value of each class can be approximated using linear combinations of Gaussian distributions assuming all features are independent. We can approximate a new Gaussian distribution as follows:

\begin{equation}
     \mathbf{p}\left(c_{i} | \mathbf{x}\right) \sim \mathcal{N}\left(\Sigma_{j}^{R} a_{j} \mu_{j}, \Sigma_{j}^{D} a_{j}^{2} \sigma_{j}^{2}\right),
         \label{E7}
\end{equation}
where $a$ is the weights connected to the corresponding classification node. Note that if there are more than one fully connected layer, the same process are performed iteratively. We can use the variance measure to either reject the input or concern the reliability of the model's predictions. We also note that these measures act like confidence for the predictions.

\begin{figure*}[t]
    \begin{center}
    \includegraphics[width=1.0\linewidth]{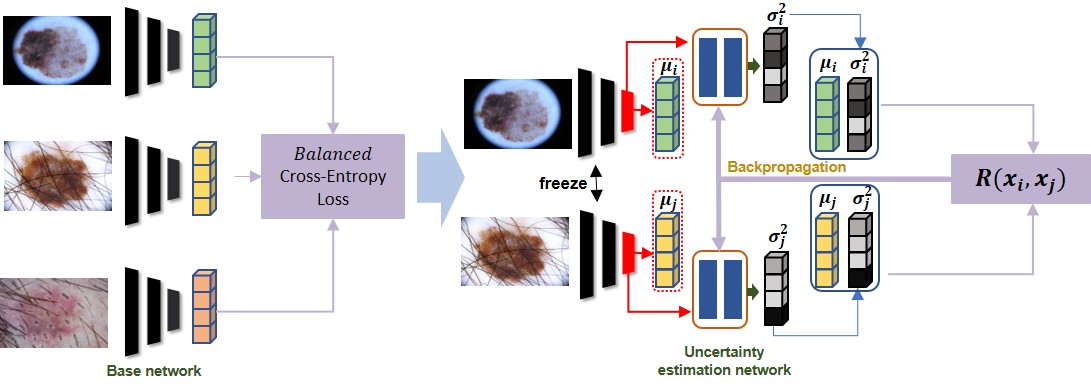}
    \end{center}
    \caption{The training procedure of our network. The base network is trained to estimate the $\mu$ value and the confidence network is trained to estimate the $\sigma^{2}$ value for the corresponding feature.}
    \label{fig:fig3}
\end{figure*}

\subsection{Training}

Our work focuses on estimating a probabilistic Gaussian distribution of the features instead of discrete features. In order to do so, we estimate the $\mu$ and $\sigma^{2}$ value for
each corresponding feature. We can train the $\mu$ by training the base network as any other deterministic model does. We choose the backbone network and train the $\mu$ value with the image and corresponding label. Using the same bottleneck layer, which is the layer before concatenation of $\mu$, we use that layer as an input to the confidence network which will output initial variance value at first. This will be optimized along with the training process.

We use a two-stage training strategy. First, training the backbone network with training dataset using weighted cross-entropy loss. Second, given a pretrained model $f$, we fix its parameters, take $\mu(\mathrm{x})=\mathrm{f}(\mathrm{x})$, and optimize an additional uncertainty estimating network to estimate $\sigma(x)^{2}$. The uncertainty estimation network shares the corresponding layer in order to estimate the variance for each latent feature value. We form a loss function using the Eq. \ref{E4} for all genuine pairs $(x_{i}, x_{j})$. Formally, the loss function to minimize is as follows:

\begin{equation}
    \mathrm{L}=\frac{1}{|P|} \sum_{(i, j) \in P}^{D}-\mathrm{R}\left(z_{i}, z_{j}\right),
    \label{E8}
\end{equation}
where $P$ is the set of all genuine pairs. In practice, the loss function is optimized within each mini-batch. Intuitively, this loss function can be known and understood as an alternative to maximize $p(z|x)$: if the latent distributions of all possible genuine pairs have a large overlap, the latent target $z$ should have a large likelihood for and corresponding input. Note that since the mean value is fixed, the optimization would not lead the $\mu$ values to collapse.

Among the two uncertainties introduced in Section 2, our work focuses on modeling epistemic uncertainty which is helpful to identify out-of-distribution images or noisy inputs.
\begin{table*}[h]
\begin{center}
\caption{Results for ISIC 2018 with each backbone network. After we tested on various base models, we added the uncertainty estimation network for confidence pooling (CP). Then again, we measured the performance on F1, accuracy (ACC), balanced accuracy (BACC), and mean AUC.}
\begin{tabular}{|p{3cm}|p{1.5cm}|p{1.5cm}|p{1.5cm}|p{2cm}|}
\hline
Model             & F1               & ACC.     & BACC.                   & Mean AUC          \\ \hline\hline
ResNet50          & 63.5\% & 83.3\% & 67.5\% & 89.6\% \\
ResNet101         & 63.3\% & 84.6\% & 67.5\% & 88.9\%  \\
Densenet121       & 64.1\% & 85.0\% & 67.0\% & 87.6\%  \\
Densenet201       & 64.5\%  & 86.5\% & 68.7\% & 88.6\%  \\
se-Resnext50      & 65.2\% & 87.1\% & 70.4\% & 90.3\%  \\
se-Resnext101     & 66.5\%  & 88.7\% & 69.9\% & 86.3\%  \\
Senet154          & 66.3\% & 86.2\% & 68.1\% & 87.2\%  \\
EfficientNetb3    & 68.2\% & 89.8\% & 73.4\% & 93.7\% \\
EfficientNetb4    & 69.0\% & 90.7\% & 74.2\% & 94.1\% \\ \hline
ResNet50+CP       & 65.4\% & 85.5\% & 71.1\% & 92.1\% \\
ResNet101+CP      & 65.1\% & 86.2\%  & 71.2\% & 90.6\% \\
Densenet121+CP    & 66.5\% & 87.1\% & 71.9\% & 89.0\% \\
Densenet201+CP    & 66.9\% & 88.2\% & 72.2\% & 89.1\%  \\
se-Resnext50+CP   & 67.0\%  & 89.5\% & 73.1\% & 92.1\%  \\ 
se-Resnext101+CP  & 68.1\%  & 90.9\% & 74.0\% & 89.9\%  \\ 
Senet154+CP       & 66.5\% & 86.5\% & 70.2\% & 87.0\% \\
EfficientNetb3+CP & 70.1\% & 91.1\% & 77.0\% & 95.5\%  \\
EfficientNetb4+CP & 71.2\% & 92.3\% & 78.7\% & 95.7\%  \\ \hline
\end{tabular}
\label{T1}
\end{center}
\end{table*}
\section{Dataset and Experiment}

Our work is evaluated on the publicly available ISIC 2018 \cite{Dat1} and ISIC 2019 \cite{Dat1,Dat2,Dat3}. ISIC 2018 or HAM10000 dataset contains 10,015 images with 7 classes, including melanoma(MEL), melanocytic nevus(NV), basal cell carcinoma(BCC), Actinic keratosis(AK), benign keratosis(BK), dermatofibroma(DF) and vascular lesion(VS). The largest category, melanocytic nevus has 6,705 images while the smallest category, dermatofibroma only has 115 images. All of the images were preprocessed in order to normalize their intensity.

We also carried out an experiment on the ISIC 2019 set, which comprises of 25,331 dermoscopic images. The following dataset contains 8 classes, including melanoma, melanocytic nevus, basal cell carcinoma, actinic keratosis, benign keratosis, dermatofibroma, vascular lesion and squamous cell carcinoma(SCC). We do not augment the training set with external data, and we randomly split the available data into training (80\%) and test (20\%) sets. We use 5-fold cross validation for evaluation.

\subsection{Implementation Details}
We conduct various experiments with different deep architectures, including ResNet50, ResNet101, Densenet121, Densenet201, se-Resnext50, se-Resnext101, Senet154, EfficientNetb3 and EfficientNetb4. Before training the network, we resize the input image differently for each backbone network. We mainly resize the image to $256 \times 256$ pixels and use normalization within the range of [0, 1]. For base model training, we choose Adam optimization \cite{adam} with a batch-size of 32 as the model parameter. We initialize the learning rate at 0.01 and decay it by 0.1 every 50 epochs. The uncertainty estimation network is composed of two fully connected layers with 1,024 dimensions. We use Adam optimization with a batch size of 32 with the learning rate of 0.005. We decay the learning rate by 0.1 every 100 epochs. Our algorithm runs on two NVIDIA TITAN Xp GPUs with 12 GB of VRam.

\begin{table*}[]
\begin{center}
\caption{Results for ISIC 2019 with each backbone network. We tested on two settings, with and without uncertainty estimation, which is used for confidence pooling. We measure performance on F1, accuracy, balanced accuracy, and mean AUC.}
\begin{tabular}{|p{3cm}|p{1.5cm}|p{1.5cm}|p{1.5cm}|p{2cm}|}
\hline
Model             & F1               & ACC.              & BACC.     & mean AUC          \\ \hline\hline

Densenet201       & 59.5\% & 72.5\% & 65.7\% & 88.6\% \\
se-Resnext101     & 60.5\% & 72.7\% & 64.9\% & 86.3\%  \\
Senet154          & 61.3\% & 68.2\% & 63.1\% & 86.2\%  \\
EfficientNetb4    & 63.0\% & 75.8\% & 71.1\% & 89.1\% \\ \hline
Densenet201+CP    & 62.9\% & 74.2\% & 69.2\% & 89.7\%  \\
se-Resnext101+CP  & 63.1\% & 75.9\% & 70.0\% & 89.9\% \\
Senet154+CP       & 65.5\% & 71.3\% & 68.5\%   & 88.8\% \\
EfficientNetb4+CP & 67.2\% & 79.0\% & 75.5\%  & 91.9\% \\ \hline
\end{tabular}
\label{T4}
\end{center}
\end{table*}

\begin{table*}[h]
\begin{center}
\caption{Results for each class on EfficientNetb4 with uncertainty estimation. We compare the results between two models with and without uncertainty estimation. We evaluate all categories with sensitivity, specificity and accuracy. Note that accuracy here is not the precision.}
\begin{tabular}{|c|c|c|c|c|c|c|}
\hline
    & Sensitivity & Specificity &  ACC. & Sensitivity (Ours)& Specificity (Ours) & ACC. (Ours) \\ \hline\hline
Mel & 0.76 & 0.91 & 0.89       & 0.80 & 0.92 & 0.90          \\
NV  & 0.81 & 0.95 & 0.88  & 0.82 & 0.97 & 0.90          \\
BCC & 0.77 & 0.96 &0.94       & 0.85  & 0.96  & 0.95          \\
AK  & 0.70 & 0.96 & 0.95      & 0.70 & 0.97 & 0.96          \\
BK  & 0.56 & 0.98 & 0.94      & 0.62 & 0.98 & 0.94          \\
DF  & 0.68 & 0.99 & 0.98       & 0.78 & 0.99 & 0.99          \\
VS  & 0.76 & 0.98 & 0.98       & 0.80 & 0.98 & 0.98          \\
SCC & 0.66 & 0.97 & 0.96      & 0.67 & 0.98 & 0.97          \\ \hline
\end{tabular}
\label{T6}
\end{center}
\end{table*}

\begin{figure*}[h]
    \begin{center}
    \includegraphics[scale=0.7]{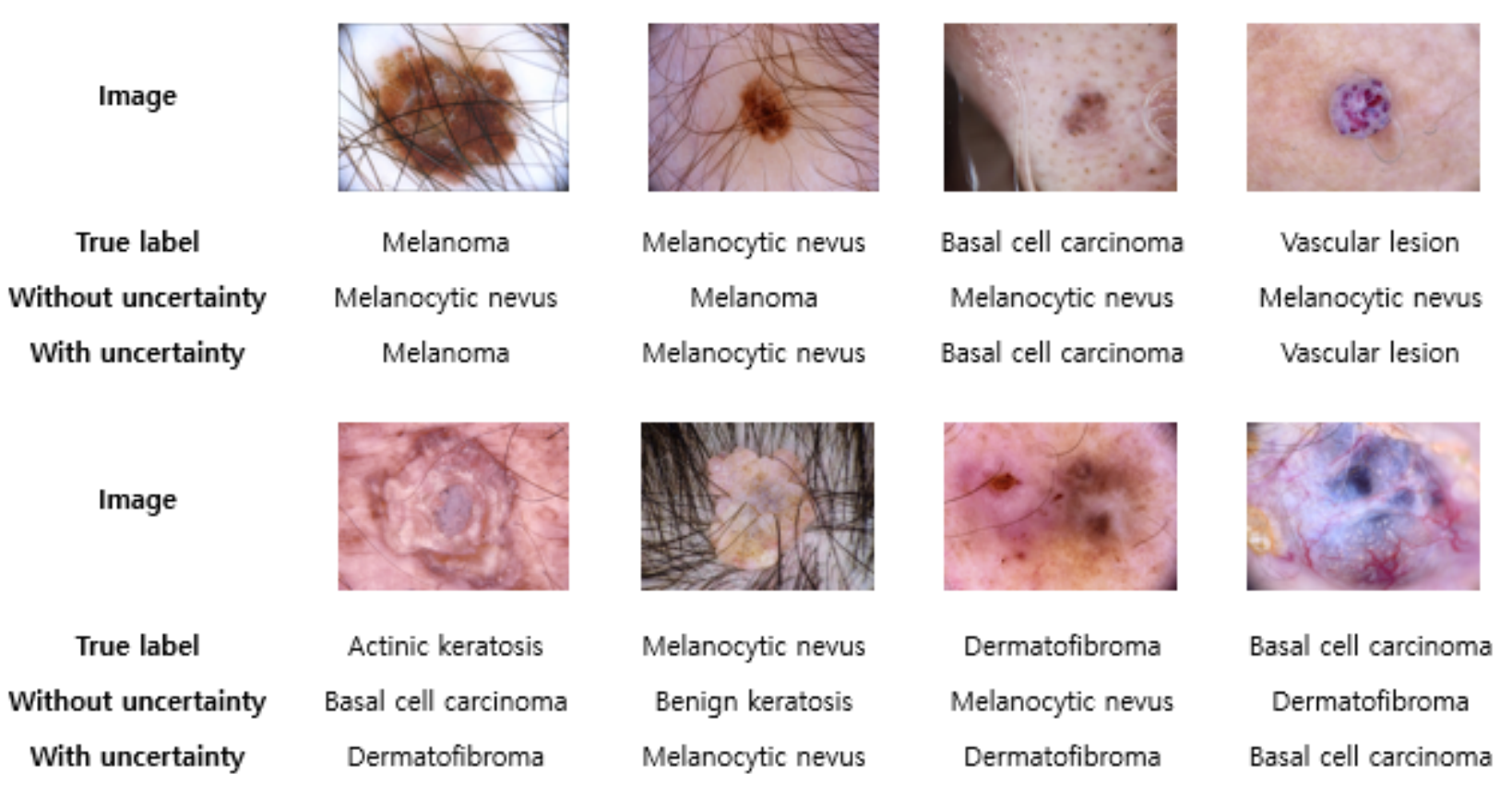}
    \end{center}
    \caption{Ablation study on confidence pooling. Image used for the second row are the images which are selected for high uncertainty. We sorted test images for each category with their confidence value and selected images within 10\% low confidence.}
    \label{fig:fig4}
\end{figure*}

\begin{table}[h]
\begin{center}
\caption{Experiments using EfficientNetb4 to investigate the benefit of rejecting uncertain inputs based on the uncertainty measure. Here, we sort out images with the highest uncertainty for each category in the test set of ISIC 2019.}
\begin{tabular}{|p{2cm}|p{1cm}|p{1cm}|p{1.5cm}|}
\hline
  Rejection Ratio       & F1      & ACC      & BACC \\ \hline\hline
\textbf{Ours [0\%]}  & 67.2 \% & 79.0 \%  & 75.5 \%     \\
\textbf{Ours [5\%]} & 71.1 \% & 86.9 \% & 82.1 \%     \\
\textbf{Ours [10\%]} & 74.2 \% & 89.2 \%  & 85.3 \%     \\
\textbf{Ours [20\%]} & 75.5 \% & 90.7 \%  & 86.7 \%     \\ \hline
\end{tabular}
\label{T5}
\end{center}
\end{table}
\section{Results}

We evaluated the proposed model on a test set of dermoscopy images and quantified its performance using various metrics regarding the class-specific sensitivity (SE) and specificity (SP), and the overall balanced accuracy (BACC). The balanced multi-class accuracy is defined as the measure of accuracy of each category weighted by the category prevalence. Specifically, it can be interpreted as the arithmetic mean of the (true positives / positives) across each categories of diseases. Note that this metric is semantically similar to the average recall score. The Area Under the Curve (AUC) for each class was also computed. We carried an experiment on only using base architectures, however, we can easily add the uncertainty estimating network to any other existing models.

\subsection{ISIC 2018} \label{rrr}

Table \ref{T1} shows the performance for each network with and without uncertainty information on ISIC 2018 test set. We tested our confidence network with using various backbone architectures; ResNet, Densenet, se-Resnext, Senet, and EfficientNet. Note that we choosed the backbone architectures which are frequently used for the competitions. As shown in Table \ref{T1}, we could clearly see that concerning uncertainties actually improves the performances of all metrics for almost all architectures. The accuarcy had about 2 \% improvement while the balanced accuracy had improvement about 4 \%. This shows that by pooling the confidence to the latent features, the additional uncertainty information actually assists the decision layers to focus more on confident features, therefore bringing improvements to the model.

\subsection{ISIC 2019}

We perform more experiments on ISIC 2019 dataset due to its diverse and more advanced data compared to ISIC 2018. Table \ref{T4} shows the performance with each backbone network with and without uncertainty information on ISIC 2019 validation set. Again, we use the architectures which are frequently used in competitions along with models which performed relatively good in section \ref{rrr}. Similar to the experiment we did before, additional uncertainty information helped to improve all metrics for most models. As a result, we can be assured that estimating uncertainty and using it for classification level actually improves results by helping the decision layers to avoid uncertain features for dermatological image classification tasks.

To take a closer look, Table \ref{T6} shows the sensitivity, specificity, accuracy, and AUC for each category using EfficentNetb4 as the backbone structure. Since melanocytic nevus has bigger proportions in the dataset, it achieves higher performance. Moreover, adding uncertainty information to the decision layers again further improves all four metrics especially sensitivity, which is very important for medical predictions. Also, as shown in Figure \ref{fig:fig4}, we show the effect on applying uncertainties by showing some examples with high uncertainty. The images in the second row are modeled with very low prediction confidence. Note that these low prediction confidence can be due to the hair, color distribution or some other uncertain features which shift the features from its category's cluster. While other deterministic base models tend to misclassify these uncertain images, we can see that most uncertain data are correctly classified by the assist of confidence measures.

\noindent\textbf{Rejecting Data on Test Set.} The measured confidence for each feature can be efficiently used in the process of computer-aided diagnosis which would help the interaction between the doctor and machine. As mentioned in Section 1, it would be best if the model can share its opinion with the level of confidence for the prediction. Since our model can additionally model confidence as in section \ref{cm}, we can think of a way to reject the input with high uncertainty. For experiment, we collected the confidence of the prediction for each category. The prediction confidence for each corresponding input is sorted which then we eliminate 0\%, 5\%, 10\%, 20\% of test samples with eliminating the lowest confidence from the measurement. We use the F1 score, accuracy and balanced accuracy for performance analysis. The metrics improved on the test set as we increased the amount of rejection as in Table \ref{T5}. This is quite an increase, demonstrating the potential of this strategy to improve the robustness of the model. Without any Bayesian techniques or inference, our method can be easily applied to any base models in the needs of modeling confidence. Note that, in the real world, a doctor would set a threshold of this confidence level to automatically reject inputs.

\section{Conclusion}
This paper proposes a dermatological lesion classification network with uncertainty estimation. Our qualitative experiments show that these measures of uncertainty can not only be used for modeling confidence for the output but also increase the performance for classification tasks. We evaluated our method on the ISIC datasets and show that our strategy has the potential of having wide applications either to improve the performance of the model. Moreover, the obtained uncertainty can be used to reject inputs with low confidence or actively ask and interact with the expert, as well as to provide more insightful information in the diagnosis process.
%
%
%
%

\end{document}